\def\year{2019}\relax
\documentclass[letterpaper]{article} 
\usepackage{aaai19}  
\usepackage{times}  
\usepackage{helvet}  
\usepackage{courier}  
\usepackage{url}  
\usepackage{graphicx}  
\usepackage{subfig}
\usepackage{amsfonts,amssymb}
\usepackage{url}
\begin{document}
%
\title{Spatial-Temporal Person Re-identification}
\author{Guangcong Wang$^{1}$, Jianhuang Lai$^{1,2,3}\thanks{Corresponding author}$, Peigen Huang$^{1}$, Xiaohua Xie$^{1,2,3}$\\
$^{1}$School of Data and Computer Science, Sun Yat-sen University, China\\
$^{2}$Guangdong Key Laboratory of Information Security Technology\\
$^{3}$Key Laboratory of Machine Intelligence and Advanced Computing, Ministry of Education\\
\{wanggc3, huangpg\}@mail2.sysu.edu.cn, \{stsljh, xiexiaoh6\}@mail.sysu.edu.cn\\
}
\maketitle
\begin{abstract}
Most of current person re-identification (ReID) methods neglect a spatial-temporal constraint. Given a query image, conventional methods compute the feature distances between the query image and all the gallery images and return a similarity ranked table. When the gallery database is very large in practice, these approaches fail to obtain a good performance due to appearance ambiguity across different camera views. In this paper, we propose a novel two-stream spatial-temporal person ReID (st-ReID) framework that mines both visual semantic information and spatial-temporal information. To this end, a joint similarity metric with Logistic Smoothing (LS) is introduced to integrate two kinds of heterogeneous information into a unified framework. To approximate a complex spatial-temporal probability distribution, we develop a fast Histogram-Parzen (HP) method. With the help of the spatial-temporal constraint, the st-ReID model eliminates lots of irrelevant images and thus narrows the gallery database. Without bells and whistles, our st-ReID method achieves rank-1 accuracy of 98.1\% on Market-1501 and 94.4\% on DukeMTMC-reID, improving from the baselines 91.2\% and 83.8\%, respectively, outperforming all previous state-of-the-art methods by a large margin. Code is available at \url{https://github.com/Wanggcong/Spatial-Temporal-Re-identification}.
\end{abstract}

\section{Introduction}
\noindent Person ReID aims to re-target pedestrian images across non-overlapping camera views given a query image. Recently, state-of-the-art person ReID methods \cite{Wang2016DARI,Zhong2017CVPR,Bai2017CVPR,Tang2017CVPR,zhuo2018occluded,lin2016cross} gained a significant improvement (e.g., rank-1 accuracy of 80-90\% on Market-1501) by using deep learning for feature representation. However, these methods are still far from applied in real-world scenarios that may contain a large amount of gallery images. It is hard to further improve the performance using only general visual features due to appearance ambiguity. For example, different persons may share a similar appearance, a lighting condition or a human pose. How to exploit extra information to get around this bottleneck becomes a hot topic in person ReID community.

Recent studies attempt to exploit person structure information to improve the performance of ReID methods. They believe that person structure information, such as body parts, human poses, person attributes, and background context information, can help ReID methods capture discriminative local visual features. For example, part-based methods \cite{Li_2017_CVPR,Zhao_2017_ICCV} make a assumption that a person image consists of head, upper body, lower body and foot from top to bottom. Considering the person structure information, they can jointly learn both the global full-body and local body-part features for person ReID. Pose-based methods \cite{Su2017ICCV,Zhao_2017_CVPR} aim to extract pose-invariant features by exploiting keypoint annotations to localize and align the poses. Other methods mine the cues of attribute, semantic segmentation or background context \cite{Kalayeh_2018_CVPR,Song_2018_CVPR} for person ReID. However, these models obtain a limited improvement for person ReID to address the appearance ambiguity problem.
\begin{figure}[!t]
\centering
\includegraphics[width=3in]{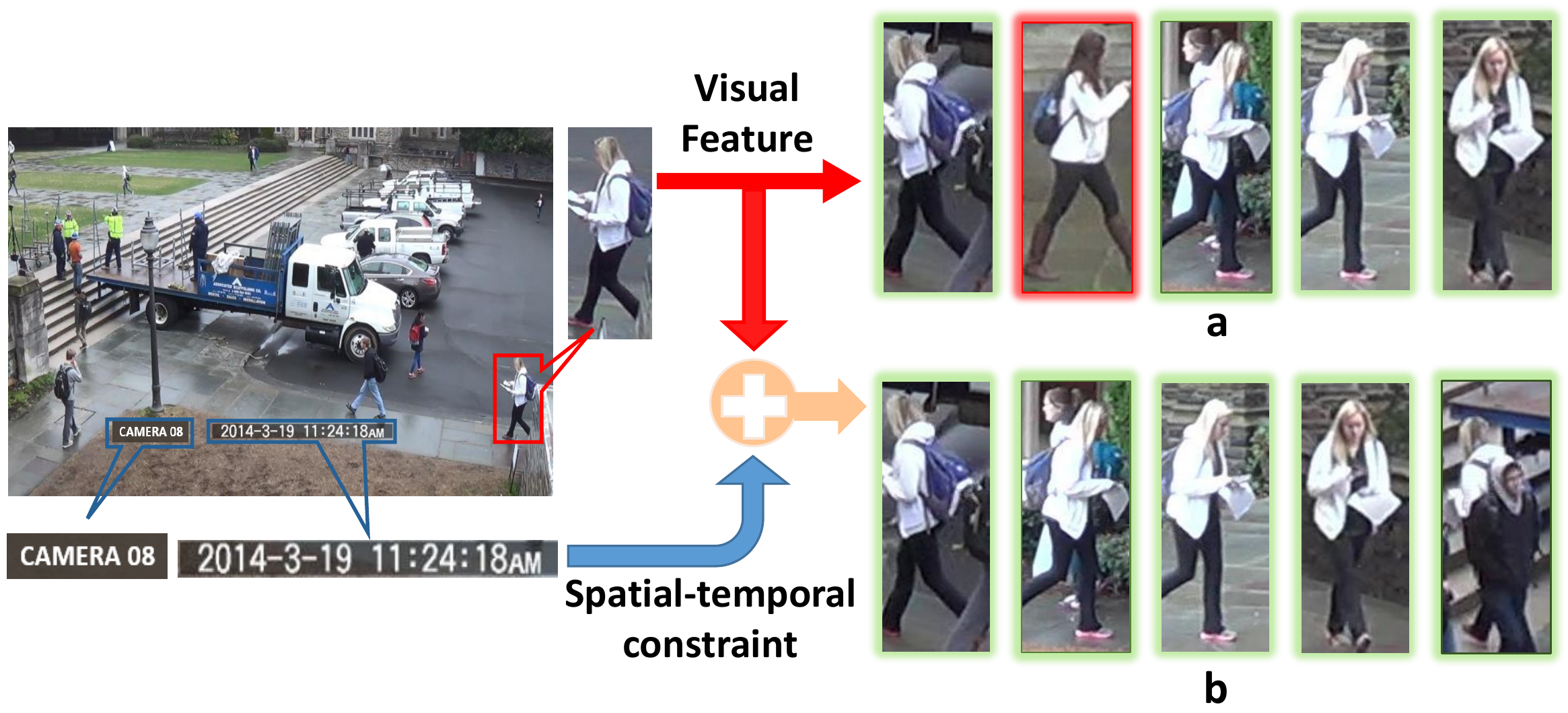}
\caption{Conventional person ReID vs. our st-ReID. (a) Retrieval results of the conventional person ReID. Without the help of spatial-temporal information, it is difficult for the conventional ReID to deal with the appearance ambiguity (red boxes denote false alarms). (b) Retrieval results of our st-ReID. With spatial-temporal information, st-ReID can eliminate irrelevant images. Besides, spatial-temporal information (camera ID and timestamp) of st-ReID widely exists in video surveillance and can be easily collected without any manual annotation. (Best viewed in color)}
\label{fig:define}
\end{figure}

Instead of using person structure information, a wide variety of approaches also attempt to exploit spatial-temporal information. The straightforward way is to exploit both spatial and temporal information from videos. Image-to-video and video-based person ReID methods \cite{wang2017tcsvt,Li_2018_CVPR} aim to learn spatial- and temporal-invariant visual features. However, these approaches still focus on visual feature representations, but not a spatial-temporal constraint across different cameras. For example, a person captured by Camera 1 at $t$ should not be captured by Camera 2 that is far away from Camera 1 at $t+\Delta t$ ($\Delta t$ is a small value). Such a spatial-temporal constraint eliminates lots of irrelevant target images in gallery, and thus significantly alleviates the appearance ambiguity problem. To distinguish the spatial-temporal concept of video-based methods, we call this the spatial-temporal person ReID (st-ReID), as shown in Figure \ref{fig:define}.
\begin{figure*}[!t]
\centering
\includegraphics[width=6in]{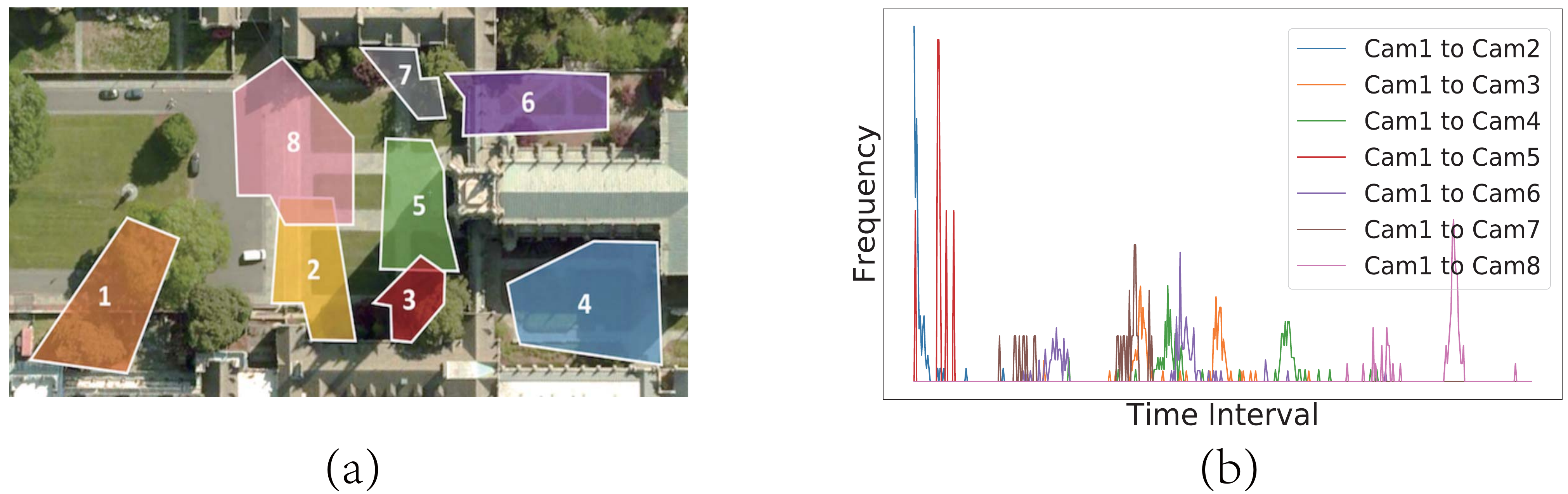}
\caption{(a) Camera topology of DukeMTMC-reID. (b) Spatial-temporal distribution, i.e, frequency of positive image pairs (an image pair with the same person identity denotes a positive pair) with respect to time interval. (Best viewed in color)}
\label{fig:distribution}
\end{figure*}

St-ReID is explicitly or implicitly investigated in distributed camera network topology inference \cite{huang2016mmm,yeong} and cross-camera multiple object tracking. However, these approaches either make some strong assumptions for model simplification or do not focus on how to build an effective joint metric for the visual similarity and the spatial-temporal distribution. Formally, st-ReID is to learn a mapping $f:X\rightarrow Y$ from a training set $\{({\bf{x}}^v_i,x^s_i,x^t_i,y_i)\}$, where ${\bf{x}}^v_i$, $x^s_i$, $x^t_i$ and $y_i$ represent a visual feature vector, a camera ID (spatial information), a timestamp, and a person ID, respectively. St-ReID has three properties: 1) extra information of st-ReID (i.e., $x^s_i$,$x^t_i$) widely exists in video surveillance and can be easily collected without any manual annotation (see Figure \ref{fig:define}); 2) with the cheap spatial-temporal information, the performance of ReID can be significantly improved (6.9\% and 10.6\% improvement on Market1501 and DukeMTMC-reID, respectively); 3) st-ReID can be thought of as analogous to the more difficult version of cross-camera multiple object tracking, which misses lots of in-between frames. St-ReID bridges the gap between the conventional person ReID and the cross-camera multiple object tracking.

There are three key challenges to model the spatial-temporal pattern in person ReID. First, it is extremely difficult to estimate the spatial-temporal pattern of person ReID that follows a complex distribution. Take the  DukeMTMC-reID dataset as an example (Figure \ref{fig:distribution} (a)), there are several paths between Camera 1 and Camera 6 and therefore several peaks exist in the spatial-temporal distribution from Camera 1 to Camera 6 (Figure \ref{fig:distribution} (b)). Second, even though we can find a good formulation to describe the complex spatial-temporal distribution based on a finite dataset, it is still unreliable due to uncertain walking trajectories and velocities. That is, a person may appear at anytime and from anywhere. Third, given a reliable visual appearance similarity and an unreliable spatial-temporal distribution, it is difficult to build a reliable joint metric because the spatial-temporal distribution is unreliable and it is hard to assign appropriate weighting factors for these two types of metrics.

Considering these intractable problems, a novel joint similarity metric with Logistic Smoothing (LS) is proposed to integrate both visual feature similarity and spatial-temporal patterns into a unified metric function. Specially, we first train a deep convolutional neural network for visual feature representation based on the PCB model \cite{Yifan2017arxiv}. A fast Histogram-Parzen method (HP) is then introduced to describe the probability of positive image pairs with respect to time difference for each camera pair. To avoid missing low-probability positive gallery images, we propose to use logistic smoothing (LS) to alleviate the problem of uncertain walking trajectories and velocities in person ReID.

Overall, this paper makes three main contributions:
\begin{itemize}
  \item First, we propose a novel two-stream spatial-temporal person ReID (st-ReID) framework that takes both visual semantic information and spatial-temporal information into consideration. With the help of the cheap spatial-temporal information that can be easily collected without any manual annotation, the st-ReID model eliminates lots of irrelevant images and thus to alleviate the problem of appearance ambiguity in person ReID.

  \item Second, we propose a joint similarity metric with Logistic Smoothing (LS) to integrate two kinds of heterogeneous information into a unified framework. Furthermore, we develop a fast Histogram-Parzen (HP) method to approximate the spatial-temporal probability distribution.

  \item Third, without bells and whistles, our st-ReID method achieves rank-1 accuracy of 98.1\% on Market-1501 and 94.4\% on DukeMTMC-reID, improving from the baselines 91.2\% and 83.8\%, respectively, outperforming all previous state-of-the-art methods by a large margin.
\end{itemize}

\section{Related Work}
Recent person ReID methods concentrate on deep learning for visual feature representation. Basically, these deep models either attempt to design effective convolutional neural networks or adopt different kinds of loss functions, e.g., classification loss \cite{zheng2016mars,feng2018learning,Liang2018}, verification loss \cite{li2014deepreid,chen2015deep}, and triplet loss \cite{ding2015deep,wang2017tcsvt,HermansBL17,Wang2016DARI}. Due to the remarkable ability of CNN representation, state-of-the-art approaches achieve a good performance, e.g., rank-1 accuracy of 80-90\% on Market-1501. However, these methods can hardly address the appearance ambiguity problem.

In order to achieve this goal, many studies try to exploit person structure information \cite{Li_2017_CVPR,Zhao_2017_ICCV,Su2017ICCV,Zhao_2017_CVPR,Kalayeh_2018_CVPR,Song_2018_CVPR}. For example, a multi-scale context-aware network \cite{Li_2017_CVPR} is used to learn powerful features over full body and body parts to capture the local context information. A pose-driven deep convolutional model \cite{Su2017ICCV} is introduced to alleviate the pose variations and learn robust feature representations from both the global images and different local parts. A human parsing method \cite{Song_2018_CVPR} is adopted to improve the performance of person ReID with the help of the pixel-level accuracy.
\begin{figure*}[!t]
\centering
\includegraphics[width=5in]{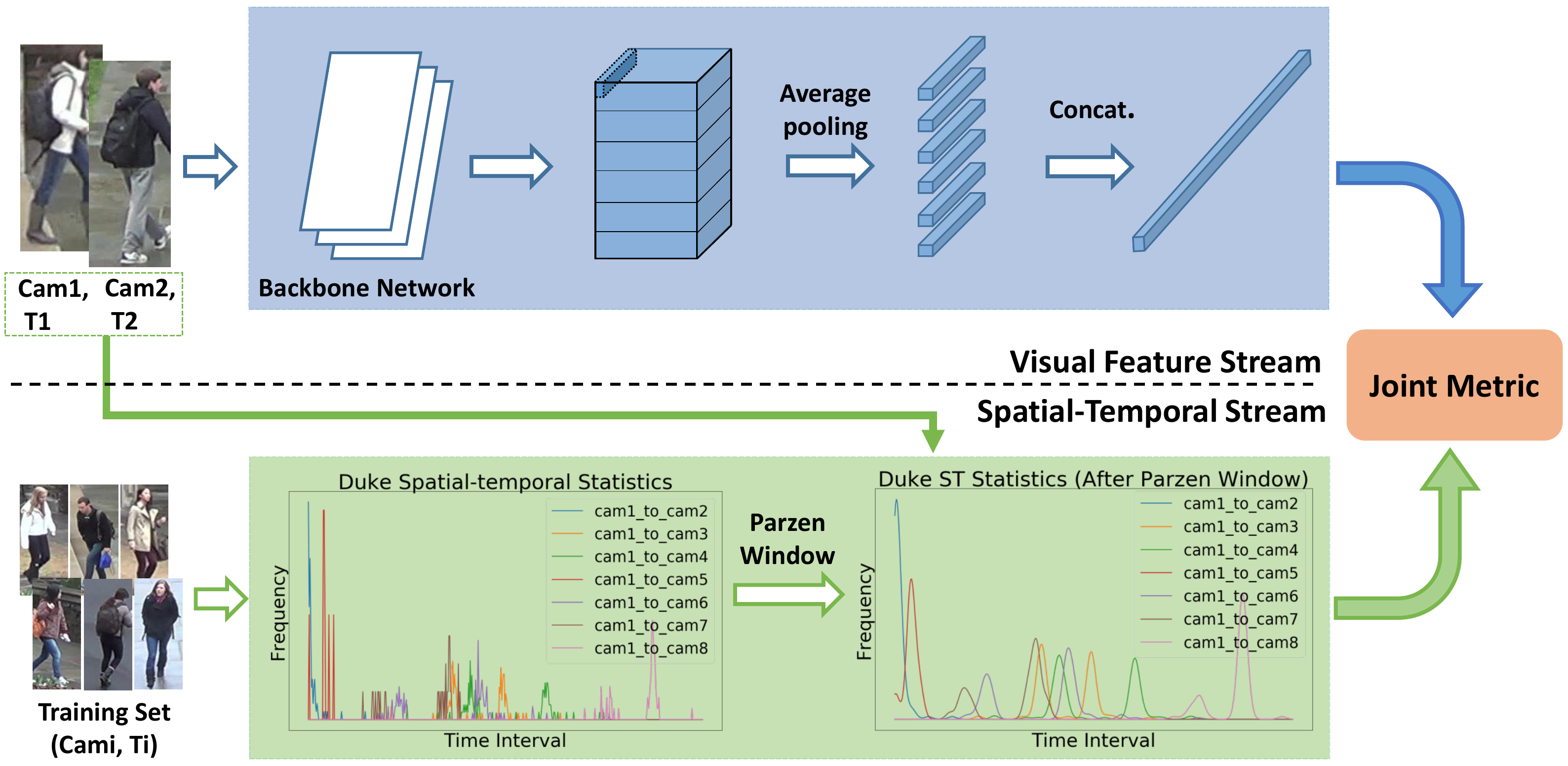}
\caption{The proposed two-stream architecture. It consists of three sub-modules, i.e., a visual feature stream, a spatial-temporal stream, and a joint metric sub-module. (Best viewed in color)}
\label{fig:overview}
\end{figure*}

Rather than using the person structure information, another group of researchers pay attention to spatial-temporal information. According to different kinds of annotations, spatial-temporal methods can be categorized into two sub-groups. In the first sub-group, spatial-temporal information is implicitly hidden in videos, e.g., image-to-video \cite{wang2017tcsvt} and video-based person ReID \cite{Li_2018_CVPR,zheng2016mars}. For example, Wang et al. \cite{wang2017tcsvt} proposed a point-to-set network for the image-to-video person ReID. Li et al. \cite{Li_2018_CVPR} introduced a spatial-temporal attention model to discover a diverse set of distinctive body parts for the video-based person ReID. In the second sub-group, spatial-temporal information is explicitly used as a constraint that eliminate the irrelevant gallery images \cite{yeong,huang2016mmm,Lv_2018_CVPR}. For example, camera network topology inference methods \cite{yeong,Lv_2018_CVPR} aim to perform the person ReID and camera network topology inference alternately in an online or unsupervised learning manner. Given a person image with a timestamp $t$, they make a strong assumption that this person should be appear at $(t-\Delta t,t+\Delta t)$. Different from these methods, our st-ReID approach seeks an effective joint metric that naturally integrates spatial-temporal information into the visual feature representation for supervised person ReID. Besides, a Camera Network based Person ReID (CNPR) \cite{huang2016mmm} is introduced to consider both the visual feature representation and the spatial-temporal constraint. However, the CNPR model makes a strong assumption that the time difference for the transition between cameras follows a Weibull distribution that contains a peak value and thus is unavailable in complex scenarios, e.g., DukeMTMC-reID. Besides, CNPR fails to address the problem of uncertain walking trajectories and velocities. Different from the CNPR model, we propose to use a Histogram-Parzen window method for the Probability Density Function (PDF) approximation and introduce a logistic smoothing approach to solve the uncertainty problem.

\section{Proposed Method}
\label{sec:met}
St-ReID aims to exploit both the visual feature similarity and the spatial-temporal constraint in a unified framework. To this end, we propose a two-stream architecture which consists of three sub-modules, i.e., a visual feature stream, a spatial-temporal stream, and a joint metric sub-module. Figure \ref{fig:overview} shows the two-stream architecture for the st-ReID.

\subsection{Visual Feature Stream}
Visual feature representation approaches are investigated in lots of studies. We do not focus on how to extract a discriminative and robust feature representation in this paper. Therefore, we use a clear Part-based Convolutional Baseline (PCB) \cite{Yifan2017arxiv} as a visual feature stream without considering a refined part pooling. This stream contains a ResNet backbone network, a stripe-based average pooling layer, six $1\times 1$ kernel-sized convolutional layers, six fully-connected layers, and six classifiers (Cross-Entropy loss).

During the training phase, each classifier is used to predict the class (person identity) of a given image. With the part-level feature representation learning scheme, PCB can learn local discriminative features and thus achieve competitive accuracy. During the test phase, six stripe-based features are concatenated into a column vector for the visual feature representation. In Figure \ref{fig:overview}, we only show the test phase of the visual feature stream.

Given two images $I_i$ and $I_j$ ($i$ and $j$ denote image indexes in a dataset), we extract visual features by using the PCB model and obtain two feature vectors, denoting $\bf{x_i}$ and $\bf{x_j}$, respectively. We compute a similarity score according to the cosine distance
\begin{equation}
    \label{eq:cosine}
    s({\bf{x_i}},{\bf{x_j}}) = \frac{{{\bf{x_i}} \cdot {\bf{x_j}}}}{{||{\bf{x_i}}||||{\bf{x_j}}{\rm{||}}}}
\end{equation}
\subsection{Spatial-temporal Stream}
A spatial-temporal stream is to capture spatial-temporal complementary information to assist the visual feature stream. Instead of using a closed form probability distribution function \cite{huang2016mmm} that follows a strong assumption, we estimate the spatial-temporal distribution by using a non-parameter estimation approach, i.e., Parzen Window approach. However, it costs much time to directly estimate a PDF because there are too much spatial-temporal data points.

To alleviate the expensive computation problem, we develop a Histogram-Parzen approach. That is, we first estimate spatial-temporal histograms and then use the Parzen Window method to smooth it. Let $({ID}_i,c_i,t_i)$ and $({ID}_j,c_j,t_j)$ ($t_i<t_j$) denote the identity labels, camera IDs, timestamps of two images $I_i$ and $I_j$, respectively. We create coarse spatial-temporal histograms to describe the probability of a positive image pair by
\begin{equation}
    \label{eq:hist}
    {\hat p}(y=1|k,c_i,c_j) = \frac{{n_{c_ic_j}^k}}{{\sum\limits_l {n_{c_ic_j}^l} }}
\end{equation}
where $k$ indicates the $k$th bin of a histogram, i.e., the time interval $t_j-t_i\in((k-1)\Delta t,k\Delta t)$. $n_{c_{i}c_{j}}^k$ represents the number of person image pairs whose time differences are at the $k$th bin from $c_i$ to $c_j$. $y=1$ denotes that $I_i$ and $I_j$ (i.e., ${ID}_i={ID}_j$) share the same person identity, while $y=0$ for different person identities ((i.e., ${ID}_i\neq{ID}_j$)).

With the Parzen Window method, we smooth the histogram by
\begin{equation}
    \label{eq:parzen}
    p(y=1|k,c_i,c_j) = \frac{1}{Z}\sum\limits_l {{\hat p}(y=1|l,c_i,c_j)K(l - k)}
\end{equation}
where $K(.)$ is a kernel and $Z = \sum\limits_k {p(y=1|k,c_i,c_j)}$ is a normalized factor. In this work, we use a gaussian function as a kernel $K$, namely
\begin{equation}
    \label{eq:gaussian}
    K(x) = \frac{1}{{\sqrt {2\pi}\sigma }}{e^{\frac{{{-x^2}}}{{2{\sigma ^2}}}}}
\end{equation}
\subsection{Joint Metric}
After we obtain two kinds of heterogeneous patterns, it is intuitively assumed that the visual similarity probability is independent of the spatial-temporal probability. The joint probability can be simply formulated as
\begin{equation}
    \label{eq:joint1}
    p(y=1|{\bf{x_i}},{\bf{x_j}},k,c_i,c_j) = s({\bf{x_i}},{\bf{x_j}})p(y=1|k,c_i,c_j)
\end{equation}
however, Eqn. \ref{eq:joint1} neglects two points. First, it is unreasonable to directly use the similarity score as the visual similarity probability, i.e., $p(y=1|{\bf{x_i}},{\bf{x_j}})\neq s({\bf{x_i}},{\bf{x_j}})$. Second, the spatial-temporal probability $p(y=1|k,c_i,c_j)$ is unreliable and uncontrollable because the walking trajectory and velocity of a person is uncertain, i.e., a person may appear at anytime and from anywhere. Directly using $p(y=1|k,c_i,c_j)$ as the spatial-temporal probability function leads to a lower recall rate while keeping the same precision. As an example, given a query image, one gallery image is with a 0.9 similarity score, and a 0.01 spatial-temporal probability, while another gallery image is with 0.3, 0.1. Eqn. \ref{eq:joint1} tends to return the second gallery image. Those who have low spatial-temporal probabilities may be regarded as irrelevant images. However, this is impractical in real-world scenarios, especially video surveillance systems. For example, when retrieving the images of a thief, (s)he may not be retrieved because (s)he may walk faster than common person and has a low spatial-temporal probability. So, can we transform the similarity score as the visual similarity probability or can we build a robust spatial-temporal probability? Our observation is two-fold:

\emph{Observation 1: Laplace smoothing.} Laplace smoothing is a technique which is widely used to estimate a prior probability in Naive Bayes
\begin{equation}
    \label{eq:Laplace}
    {p_\lambda }(Y = {d_k}) = \frac{{{m_k} + \lambda}}{{M + D\lambda }}
\end{equation}
where $d_k$ indicates the label of the $k$th category, $m_k$ indicates the number of the $k$th category, $M$ is the total number of examples, $D$ is the total number of categories, $\lambda$ is the smoothing parameter. As a special case, the number of categories $D$ is 2 and $\lambda=1$, we obtain
\begin{equation}
    \label{eq:Laplace2}
    {p_\lambda }(Y = {d_k}) = \frac{{{m_k} + 1}}{{M + 2}}
\end{equation}
We can see that Laplace smoothing is used to adjust the probability of rare (but not impossible) events so those probabilities are not exactly zero and zero-frequency problems are avoided. It serves as a type of shrinkage estimator, as the smoothing result will be between the empirical estimate $\frac{{{m_k}}}{M}$, and the uniform probability $\frac{{{1}}}{2}$.

\emph{Observation 2: Logistic function.} The logistic model is widely applied for the binary classification problem. Specially, it is defined as
\begin{equation}
    \label{eq:logistic}
    f(x;\lambda, \gamma) = \frac{1}{{1 + {{\lambda}}{e^{ - {\gamma}x}}}}
\end{equation}
where ${\lambda}$ and ${\gamma}$ are constant coefficients, ${\lambda}$ is a smoothing factor and ${\gamma}$ is a shrinking factor.

Observation 1 shows the basic idea of a smoothing operator to alleviate unreliable probability estimation. Observation 2 shows logistic function can be used for the binary classification problem. Based on these two observations, we propose a logistic smoothing approach that both adjusts the probability of rare events and compute the probability of two images belonging to the same ID given the certain information. We modify Eqn. (\ref{eq:joint1}) as
\begin{equation}
    \label{eq:joint2}
    {p_{joint}} = f(s;{\lambda}_0,{\gamma}_0)f({p_{st}};{\lambda}_1,{\gamma}_1)
\end{equation}
For notation simplicity, we use $p_{joint}$, $s$ and $p_{st}$ to denote $p(y=1|{\bf{x_i}},{\bf{x_j}},k,c_i,c_j)$, $s({\bf{x_i}},{\bf{x_j}})$ and $p(y=1|k,c_i,c_j)$, respectively. According to Eqn. (\ref{eq:cosine}) and (\ref{eq:parzen}), we can see that $s \in(-1,1)$ is shrunk by the logistic function like the Laplace smoothing, but not so much. Differently, $p_{st}\in (0,1)$ is truncated and lifted up largely. Even the spatial-temporal probability $p_{st}$ is close to zero, $f({p_{st}};{\lambda}_1,{\gamma}_1)\ge f(0)=\frac{{{1}}}{1+{\lambda}_1}$. With the logistic smoothing, Eqn. (\ref{eq:joint2}) is robust to rare events. This is reasonable because the spatial-temporal probability is unreliable as discussed above while visual similarity are relatively reliable. Besides, using the logistic function to transform the similarity score (spatial-temporal probability) into a binary classification probability (positive pair or negative pair) is intuitive and self-evident as described in Observation 2.

\subsection{Implementation Details}
As for the visual feature stream, we set the hyper-parameters following the PCB method \cite{Yifan2017arxiv} without considering the refined pooling scheme. The training images are augmented with horizontal flip and normalization and resized to $384\times 192$. We use SGD with a mini-batch size of 32. We train the visual feature stream for 60 epochs. The learning rate starts from 0.1 and is decayed to 0.01 after 40 epochs. The backbone model is pre-trained on ImageNet and the learning rate for all the pre-trained layers are set to 0.1$\times$  of the base learning rate. As for the spatial-temporal stream, we set the time interval $\Delta t$ to 100 frames. We set the gaussian kernel parameter $\sigma$ to 50 and use the three-sigma rule to further reduce the computation. As for the joint metric, we set ${\lambda}_0$, ${\lambda}_1$, ${\gamma}_0$ and ${\gamma}_1$ to 1, 2, 5 and 5, respectively.

\section{Experiments}
\label{sec:exp}
In this section, we evaluate our st-ReID method on two large-scale person ReID benchmark datasets, i.e., Market-1501 and DukeMTMC-reID, and show the superiority of the st-ReID model compared with other state-of-the-art methods. We then present ablation studies to reveal the benefits of each main component/factor of our method.

\textbf{\emph{Datasets.} }
The Market-1501 dataset is collected in front of a supermarket in Tsinghua University. A total of six cameras are used, including 5 high-resolution cameras, and one low-resolution camera. Overlap exists among different cameras. Overall, this dataset contains 32,668 annotated bounding boxes of 1,501 identities. Among them, 12,936 images from 751 identities are used for training, and 19,732 images from 750 identities plus distractors are used for gallery. As for query, 3,368 hand-drawn bounding boxes from 750 identities are adopted. In this open system, images of each identity are captured by at most six cameras. Each annotated identity is present in at least two cameras. Each image contains its camera id and frame num (time stamp).

DukeMTMC-reID is a subset of the DukeMTMC dataset for image-based re-identification. There are 1,404 identities appearing in more than two cameras and 408 identities (distractor ID) who appear in only one camera. Specially, 702 IDs are selected as the training set and the remaining 702 IDs are used as the testing set. In the testing set, one query image is picked for each ID in each camera and the remaining images are put in the gallery. In this way, there are 16,522 training images of 702 identities, 2,228 query images of the other 702 identities and 17,661 gallery images (702 ID + 408 distractor ID). Each image contains its camera id and frame num (time stamp).

\begin{table}
\begin{center}
\begin{tabular}{c|c|c|c|c}
\hline
Methods &R-1&R-5&R-10&mAP \\
\hline
BoW+kissme&44.4&63.9&72.2&20.8\\
KLFDA&46.5&71.1&79.9&-\\
Null Space&55.4&-&-&29.9\\
WARCA &45.2&68.1&76.0&-\\
\hline
PAN&82.8&-&-&63.4\\
SVDNet&82.3&92.3&95.2&62.1\\
HA-CNN&91.2&-&-&75.7\\
\hline
SSDAL&39.4&-&-&19.6\\
APR&84.3&93.2&95.2&64.7\\
\hline
Human Parsing&93.9&98.8&99.5&-\\
Mask-guided&83.79&-&-&74.3\\
Background&81.2&94.6&97.0&-\\
\hline
PDC &84.1&92.7&94.9&63.4\\
PSE+ECN&90.3&-&-&84.0\\
\hline
MultiScale&88.9&-&-&73.1\\
Spindle Net &76.9&91.5&94.6&-\\
Latent Parts&80.3&-&-&57.5\\
Part-Aligned&81.0&92.0&94.7&63.4\\
PCB(*) &91.2&97.0&98.2&75.8\\
\hline
TFusion-sup&73.1&86.4&90.5&-\\
\hline
\textbf{st-ReID}&97.2&\textbf{99.3}&99.5&86.7\\
\textbf{st-ReID+RE}&\textbf{98.1}&\textbf{99.3}&\textbf{99.6}&87.6\\
\textbf{st-ReID+RE+re-rank}&98.0&98.9&99.1&\textbf{95.5}\\
\hline
\end{tabular}
\end{center}
\caption{Comparison of the proposed method with the state-of-the-arts on Market-1501. The compared methods are categorized into seven groups. Group 1: handcrafted feature methods. Group 2: clear deep learning based methods. Group 3: attribute-based methods. Group 4: mask-guided methods. Group 5: part-based methods. Group 6: pose-based methods. Group 7: spatial-temporal methods. * denotes the methods that are reproduced by ourselves.}\label{tab:market1501}
\end{table}

\textbf{\emph{Evaluation Protocol.}} For each query, an algorithm computes the distances between the query image and all the gallery images and return a ranked table from small to large. Top-k accuracy is computed by checking if top-k gallery images contain the query identity. For each individual query identity, his/her gallery samples from the same camera are excluded due to the setting of cross-view matching in person ReID.

Mean average precision (mAP) is used to evaluate the overall performance. For each query, we calculate the area under the Precision-Recall curve, i.e., average precision (AP). Then, the mean value of APs of all queries, i.e., mAP, is calculated, which considers both precision and recall of an algorithm, thus providing a more comprehensive evaluation.
\begin{table}
\begin{center}
\begin{tabular}{c|c|c|c|c}
\hline
Methods &R-1&R-5&R-10&mAP \\
\hline
BoW+kissme&25.1&-&-&12.2\\
LOMO+XQDA&30.8&-&-&17.0\\
\hline
PAN&71.6&-&-&51.5\\
SVDNet&76.7&-&-&56.8\\
HA-CNN&80.5&-&-&63.8\\
\hline
APR&70.7&-&-&51.9\\
\hline
Human Parsing&84.4&91.9&93.7&71.0\\
\hline
PSE+ECN &85.2&-&-&79.8\\
\hline
MultiScale&79.2&-&-&60.6\\
PCB(*) &83.8&91.7&94.4&69.4\\
\hline
\textbf{st-ReID}&94.0&97.0&97.8&82.8\\
\textbf{st-ReID+RE}&94.4&\textbf{97.4}&\textbf{98.2}&83.9\\
\textbf{st-ReID+RE+re-rank}&\textbf{94.5}&96.8&97.1&\textbf{92.7}\\
\hline
\end{tabular}
\end{center}
\caption{Comparison of the proposed method with the state-of-the-arts on DukeMTMC-reID. The compared methods are categorized into seven groups. Group 1: handcrafted feature methods. Group 2: clear deep learning based methods. Group 3: attribute-based methods. Group 4: mask-guided methods. Group 5: part-based methods. Group 6: pose-based methods. Group 7: spatial-temporal methods. * denotes the methods that are reproduced by ourselves.}\label{tab:duke}
\end{table}

\subsection{Comparisons to the State-of-the-Art}
In this sub-section, we evaluate our st-ReID approach compared with lots of existing state-of-the-arts on two large-scale person ReID benchmark datesets to shows the superiority of the st-ReID approach.

\textbf{\emph{Evaluations on Market-1501.}} We evaluated the proposed st-ReID model against twenty existing state-of-the-art methods, which can be grouped into seven categories, i.e., 1) handcrafted feature methods including BoW+kissme \cite{zheng2015scalable}, KLFDA \cite{karanam2016arXiv}, Null Space \cite{Zhang_2016_CVPR} and WARCA \cite{Jose2016eccv}; 2) clear deep learning based methods including PAN \cite{zheng2017axkiv}, SVDNet \cite{Sun_2017_ICCV}, and HA-CNN; 3) attribute-based methods including SSDAL \cite{Su2016} and APR \cite{Lin2016}; 4) mask-guided methods including Human Parsing \cite{Kalayeh_2018_CVPR}, Mask-guided \cite{Song_2018_CVPR}, and Background \cite{Tian_2018_CVPR}; 5) part-based methods including  MultiScale \cite{chen2017cvprw}, PDC \cite{Su2017ICCV} and PSE+ECN \cite{Sarfraz_2018_CVPR}; 6) pose-based methods including Spindle Net \cite{Zhao_2017_CVPR}, Latent Parts \cite{Li_2017_CVPR}, Part-Aligned \cite{Zhao_2017_ICCV} and PCB 7) spatial-temporal methods including TFusion-sup \cite{Lv_2018_CVPR}.

Among them, attribute-based methods use person attribute annotations, mask-guided methods use the person masks or human body parsing annotations, part-based methods make the person body assumption or use body part detectors, and pose-based methods use keypoint annotations. These methods obtain a good performance compared with handcrafted feature methods and clear deep learning based methods, but they need expensive annotations and are quite time-consuming, e.g., pixel-level human parsing annotations, eighteen keypoints, and body part annotations.

Our st-ReID method uses the cheap spatial-temporal information (i.e., camera ID and timestamp) and achieves the rank-1 accuracy of \textbf{97.2\%} and mAP of \textbf{87.6\%}, outperforming all the existing state-of-the-art methods by a large margin. With random erase (RE) \cite{random_erase}, our st-ReID achieves  the rank-1 accuracy of \textbf{98.1\%} and mAP of \textbf{87.6\%}. With the re-ranking scheme \cite{Zhong_2017_CVPR}, our st-ReID obtains mAP of \textbf{95.5\%}. TFusion-sup also use the spatial-temporal constraint. but it makes a strong assumption that a gallery person always appears in $(t-\Delta t,t+\Delta t)$ when given a query image with a timestamp $t$. Such a method may be not effective in complex scenarios, especially  DukeMTMC-reID. Besides, TFusion-sup focuses on cross-dataset unsupervised learning for the person ReID by alternately iterating between learning visual feature representations and estimating spatial-temporal patterns. Therefore, TFusion-sup actually does not investigate how to estimate the spatial-temporal probability distribution and how to model the joint probability of the visual similarity and the spatial-temporal probability distribution.

\textbf{\emph{Evaluations on DukeMTMC-reID.}} DukeMTMC-reID is a new dataset and manifests itself as one of the most challenging reID datasets up to now. We compare our st-ReID method with ten state-of-the-art methods on the DukeMTMC-reID dataset. All of the competing methods are also evaluated on the Market-1501 dataset except LOMO+XQDA \cite{liao2015person}. As shown in Table \ref{tab:duke}, it is encouraging to see that our approach (without any re-ranking scheme) significantly outperforms the competing methods by a large margin, e.g., by improving the state-of-the-art rank-1 accuracy from 85.2\% to \textbf{94.0\%} and mAP from 79.8\% to \textbf{82.8\%} compared with PSE+ECN (with a re-ranking scheme). With random erase (RE), our st-ReID achieves  the rank-1 accuracy of \textbf{94.4\%} and mAP of \textbf{83.9\%}. With the re-ranking scheme, our st-ReID obtains mAP of \textbf{92.7\%}.

\textbf{\emph{Remarks.}} Without bells and whistles, our st-ReID model outperforms all of the previous state-of-the-art person ReID methods, e.g., a rank-1 accuracy of 98.1\% and 94.4\% on Market-1501 and  DukeMTMC-reID datasets, respectively. While outside the scope of this work, we expect many such techniques (e.g., refined pooling) to be applicable to ours.

\begin{figure*}[!t]
\centering
\subfloat[Effect of the ST and VIS streams.]{\includegraphics[width=1.6in,height=1.2in]{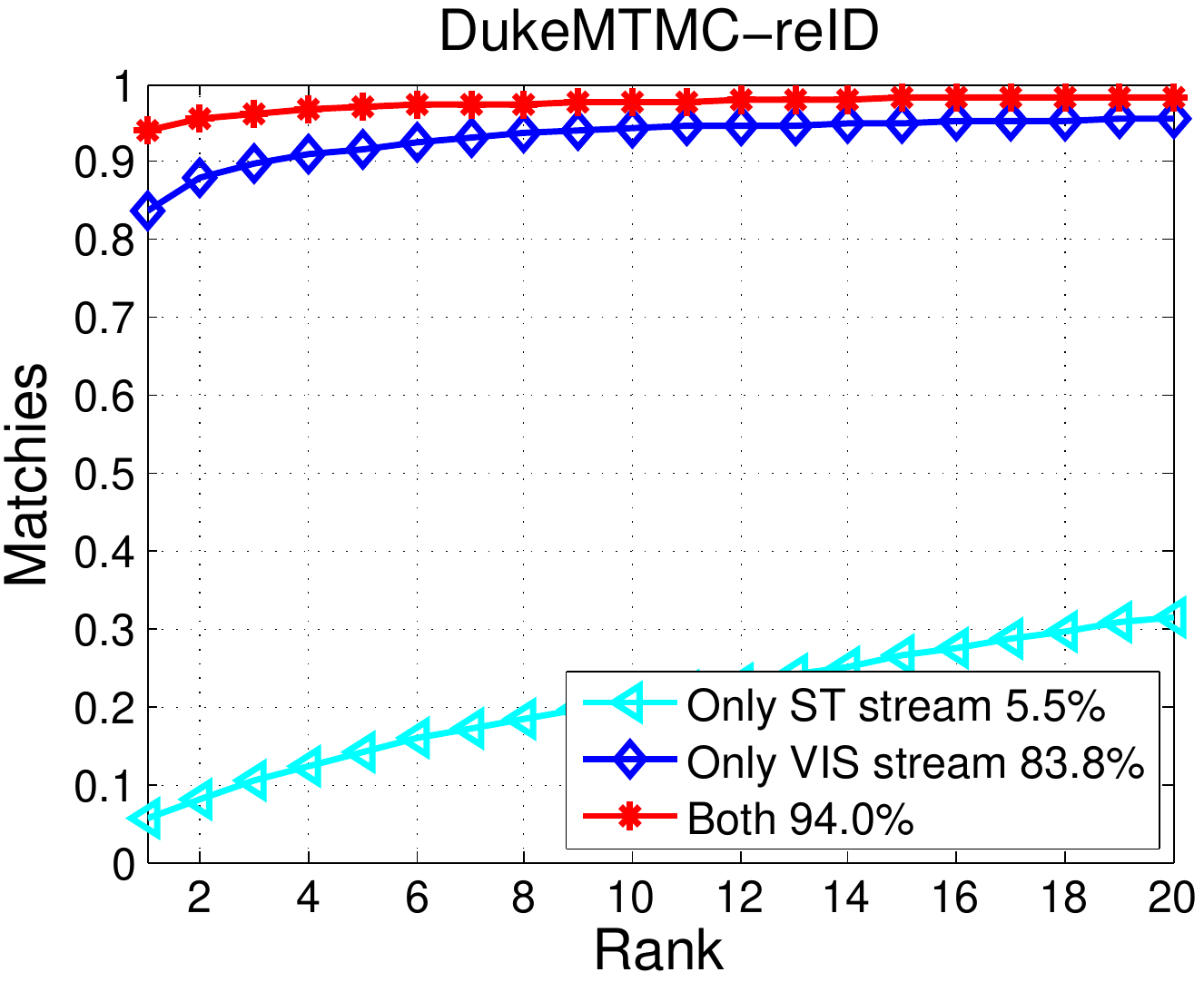}
\label{fig:stream}}
\hfil
\subfloat[Effectiveness of the joint metric.]{\includegraphics[width=1.6in,height=1.2in]{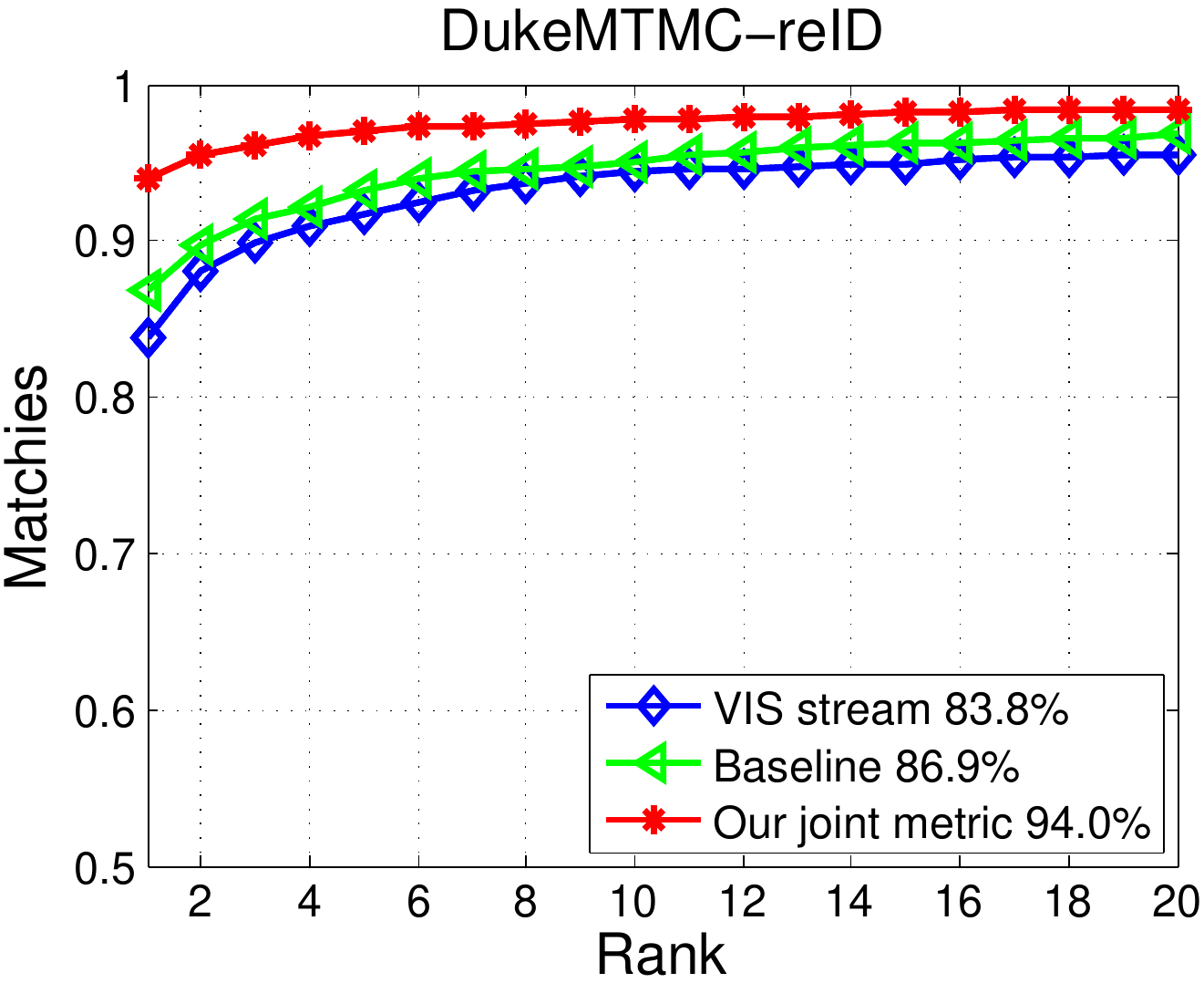}
\label{fig:joint}}
\hfil
\subfloat[Influence of $\lambda$.]{\includegraphics[width=1.6in,height=1.2in]{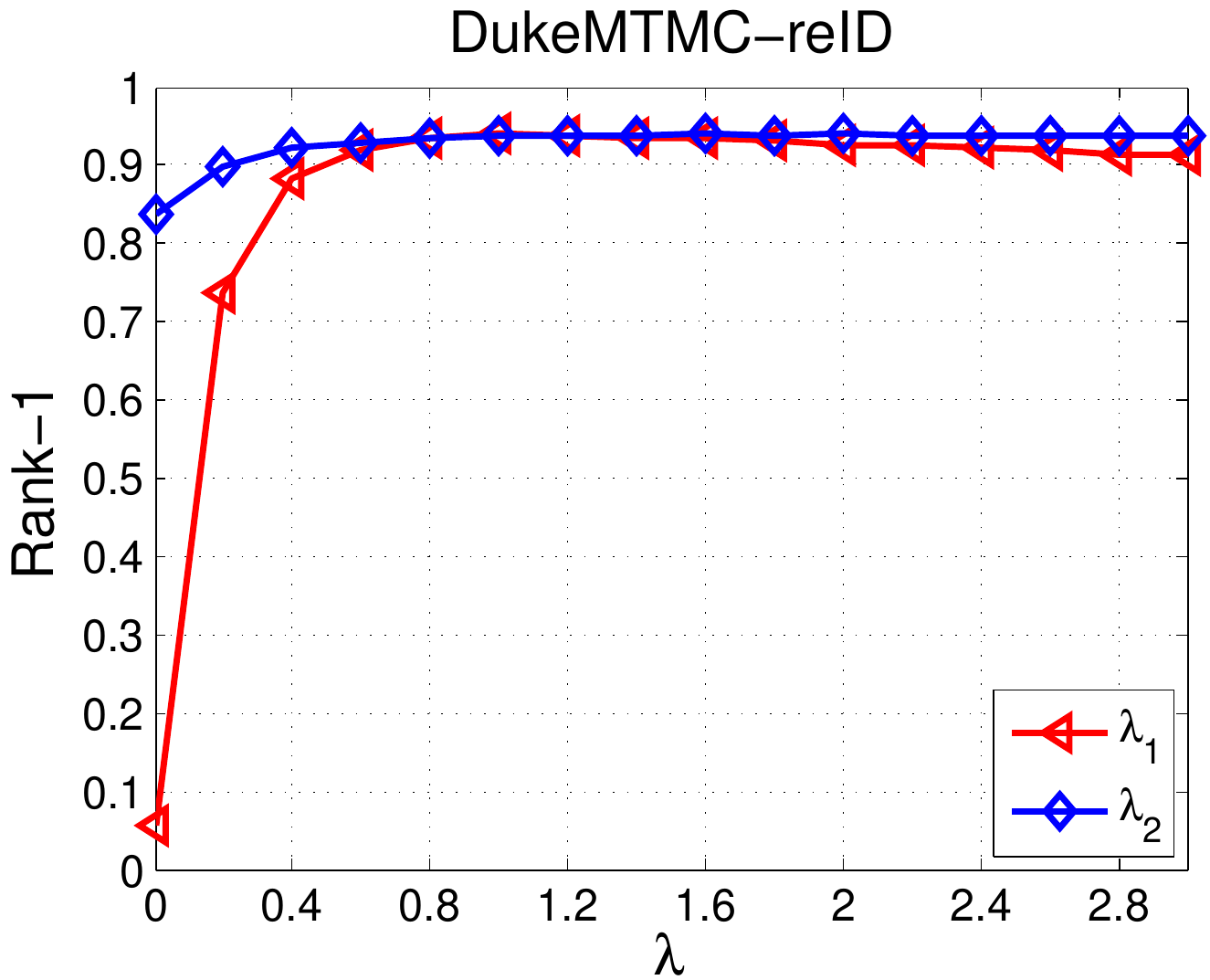}
\label{fig:fators1}}
\hfil
\subfloat[Influence of $\gamma$.]{\includegraphics[width=1.6in,height=1.2in]{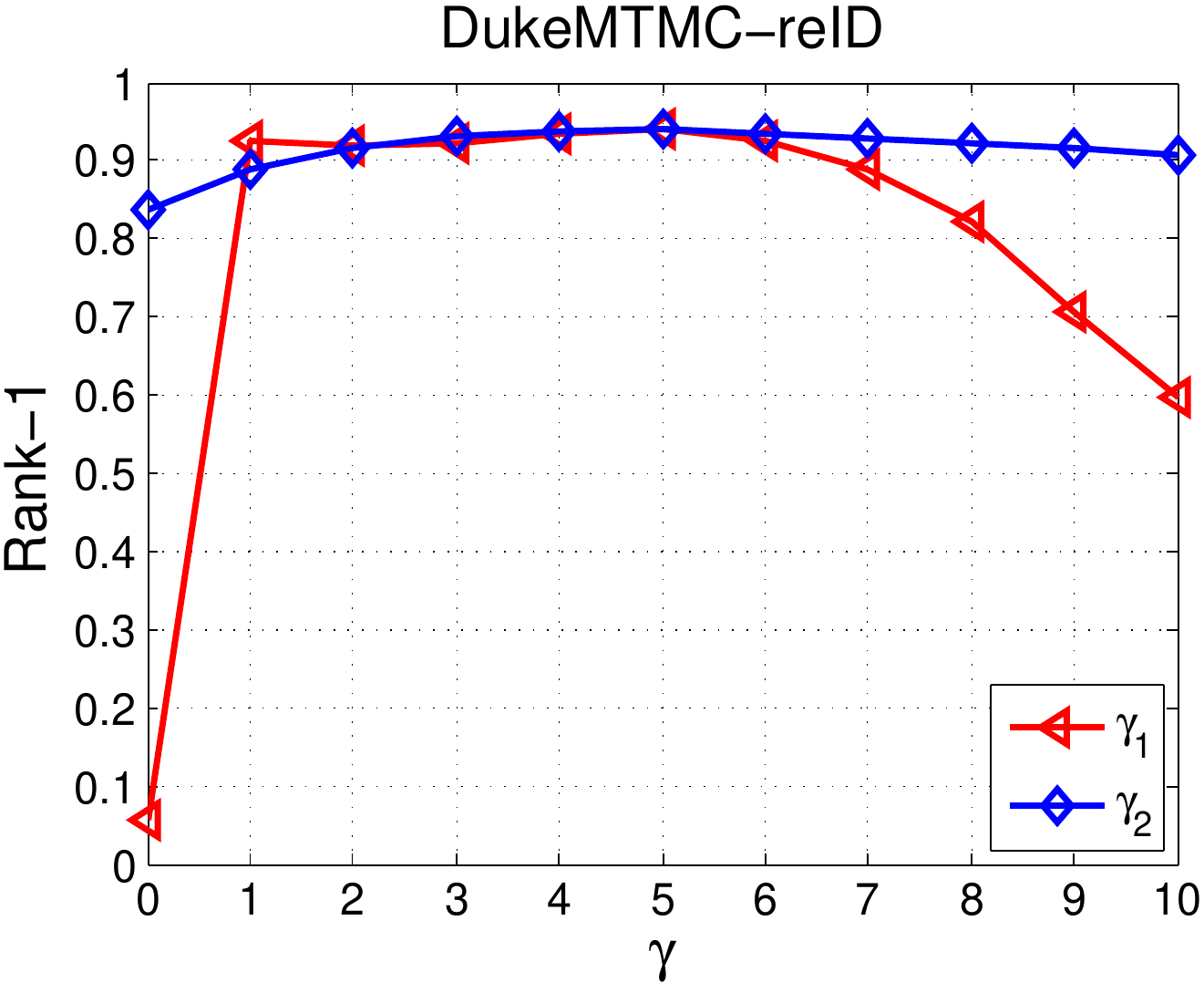}
\label{fig:fators2}}
\hfil

\caption{Effectiveness of our method.}
\label{fig:allresult}
\end{figure*}

\subsection{Ablation Studies and Model Analysis}
 To provide more insights on the performance of our approach, we conduct a lot of ablation studies on the most challenging DukeMTMC-reID dataset by isolating each key component, i.e., the visual feature representation stream, the spatial-temporal probability estimation stream and the joint metric sub-module.

\textbf{\emph{Effect of the visual feature stream.}} To show the benefit of the visual feature stream (VIS stream), we conduct an ablation study by isolating this sub-module. To achieve this, we remove the visual feature stream and thus we only use the spatial-temporal stream. It is observed that the rank-1 accuracy drops by 88.5\% (from 5.5\% to 94.0\%) when removing the VIS stream, shown in Figure \ref{fig:allresult} (a). In this experiment, we confirm that the VIS stream plays in a key role in the st-ReID approach.
\begin{table}
\begin{center}
\begin{tabular}{c|c|c|c|c}
\hline
Methods &R-1&R-5&R-10&mAP \\
\hline
ResNet50 baseline &76.9&87.8&91.0&58.7\\
\textbf{ResNet50+ST}&\textbf{87.7}&\textbf{94.1}&\textbf{95.8}&\textbf{72.2}\\
\hline
DenseNet121 baseline &79.3&89.9&92.6&63.3\\
\textbf{DenseNet121+ST}&\textbf{90.8}&\textbf{95.2}&\textbf{96.5}&\textbf{76.9}\\
\hline
PCB(*) &83.8&91.7&94.4&69.4\\
\textbf{PCB+ST}&\textbf{94.0}&\textbf{97.0}&\textbf{97.8}&\textbf{82.8}\\
\hline
\end{tabular}
\end{center}
\caption{Generalization of the st-ReID on DukeMTMC-reID.}\label{tab:generalization}
\end{table}

\textbf{\emph{Effect of the spatial-temporal stream.}} To show the benefit of the spatial-temporal stream (ST stream), we remove this sub-module to see how the spatial-temporal stream makes an effect in the st-ReID. In this case, the st-ReID model is degraded as the PCB model. As shown in \ref{fig:allresult} (a), we can see that without the spatial-temporal probability estimation stream, the rank-1 accuracy drops 10.2\% (from 94.0\% to 83.8\%).

\textbf{\emph{Effectiveness of the joint metric.}} To show the effectiveness of the joint metric, we set a baseline by using Eqn. \ref{eq:joint1}. In the baseline, both the VIS stream and the ST stream are normalized. For the fair comparison, we use the same VIS and ST streams. As shown in Figure \ref{fig:allresult} (b), our joint metric method improves the performance from 86.9\% to 94.0\%. Compared with the VIS stream (PCB model), the baseline also obtains a 3.1\% improvement because it integrates the spatial-temporal information.

\textbf{\emph{Influence of parameters.}} To investigate the impact of two important parameters in our st-ReID, i.e., the smoothing factor ${\lambda}$ and the shrinking factor ${\gamma}$, we conduct two sensitivity analysis experiments. As shown in Figure \ref{fig:allresult} (c) and (d), when ${\lambda}$ is in the range of 0.4$\sim$2.8 or ${\gamma}$ is in the range of 1$\sim$7, our model nearly keeps the best performance.

\textbf{\emph{Generalization of the st-ReID.}} To show the good generalization of the st-ReID, we further use different deep models as the VIS streams, respectively. The deep models are ResNet-50 (a clear model with the cross entropy loss), DenseNet-121 (a clear model with the cross entropy loss) and PCB. As shown in Table \ref{tab:generalization}, it is observed that when adding the ST stream into these VIS streams and using our joint metric, we can achieve more than 10\% improvement.

\section{Conclusion}
\label{sec:con}
In this paper, we propose a novel two-stream spatial-temporal person ReID (st-ReID) framework that mines both the visual semantic similarity and the spatial-temporal information. Without bells and whistles, our st-ReID method achieves rank-1 accuracy of 98.1\% on Market-1501 and 94.4\% on  DukeMTMC-reID, improving from the baseline 91.2\% and 83.8\%, respectively, outperforming all previous state-of-the-art methods by a large margin.

We intend to extend this work in two directions. First, the st-ReID builds a bridge between the conventional ReID and the cross-camera multiple object tracking and thus can be easily generalized to the cross-camera multiple object tracking. Second, we intend to further improve the performance of the st-ReID method using an end-to-end training manner.

\section{Acknowledgments}
This project was supported by the National Natural Science Foundation of China (U1611461, 61573387, 61672544).

\bibliography{egbib}
\bibliographystyle{aaai}

\end{document}